\documentclass[sn-mathphys-num]{sn-jnl}% Math and Physical Sciences Numbered Reference Style 
\usepackage{graphicx}%
\usepackage{multirow}%
\usepackage{amsmath,amssymb,amsfonts}%
\usepackage{amsthm}%
\usepackage{mathrsfs}%
\usepackage[title]{appendix}%
\usepackage{xcolor}%
\usepackage{textcomp}%
\usepackage{manyfoot}%
\usepackage{booktabs}%
\usepackage{algorithm}%
\usepackage{algorithmicx}%
\usepackage{algpseudocode}%
\usepackage{listings}%
\usepackage{xcolor}
\definecolor{light-gray}{gray}{0.95}

\newcommand{\HT}[1]{{\color{black}{#1}}}

		\newcommand{\COMMENT}[1]{}
        \newcommand{\V}[1]{ {\boldsymbol{#1}}}  %% vectors

        \newcommand{\MAT}[1] {\begin{bmatrix} #1 \end{bmatrix}}

\usepackage{cancel}

\usepackage{algorithm,setspace}
\usepackage{algpseudocode}

\usepackage{graphicx, caption}
\usepackage{soul}

 \usepackage[labelformat=simple]{subcaption}

\DeclareCaptionLabelFormat{subcaptionlabel}{\normalfont(\textbf{#2}\normalfont)}
\usepackage{amsthm}
\usepackage[normalem]{ulem}
\usepackage{tabularx}
\usepackage{svg}
\usepackage{hyperref}
\hypersetup{
    colorlinks=true,
    linkcolor=blue,
    filecolor=magenta,      
    urlcolor=cyan,
    pdftitle={Overleaf Example},
    pdfpagemode=FullScreen,
    }
\theoremstyle{thmstyleone}%
\theoremstyle{thmstyletwo}%

\theoremstyle{thmstylethree}%

\begin{document}

\title[Article Title]{Sensorized gripper for human demonstrations}

%%=============================================================%%
%% GivenName	-> \fnm{Joergen W.}
%% Particle	-> \spfx{van der} -> surname prefix
%% FamilyName	-> \sur{Ploeg}
%% Suffix	-> \sfx{IV}
%% \author*[1,2]{\fnm{Joergen W.} \spfx{van der} \sur{Ploeg} 
%%  \sfx{IV}}\email{iauthor@gmail.com}
%%=============================================================%%

\author{\fnm{Sri Harsha} \sur{Turlapati}}

\author{\fnm{Gautami} \sur{Golani}}

\author{\fnm{Mohammad Zaidi} \sur{Ariffin}}

\author*{\fnm{Domenico} \sur{Campolo}}\email{d.campolo@ntu.edu.sg}

%, Nanyang Technological University, Singapore 

\affil{\orgdiv{School of Mechanical and Aerospace Engineering}, \orgname{Nanyang Technological University}, \orgaddress{\postcode{639798}, \country{Singapore}}}

%%==================================%%
%% Sample for unstructured abstract %%
%%==================================%%

\abstract{Ease of programming is a key factor in making robots ubiquitous in unstructured environments. In this work, we present a sensorized gripper built with off-the-shelf parts, used to record human demonstrations of a box in box assembly task. With very few trials of short interval timings each, we show that a robot can repeat the task successfully. We adopt a Cartesian approach to robot motion generation by computing the joint space solution while concurrently solving for the optimal robot position, to maximise manipulability. The statistics of the human demonstration are extracted using Gaussian Mixture Models (GMM) and the robot is commanded using impedance control.}


\maketitle

\section{Towards Intuitive Robot Programming}

\begin{figure}
        \centering
  \subfloat[\textit{Tool-perspective} learning\label{FIG:big_picture}]{%
       \includegraphics[width=0.45\textwidth]{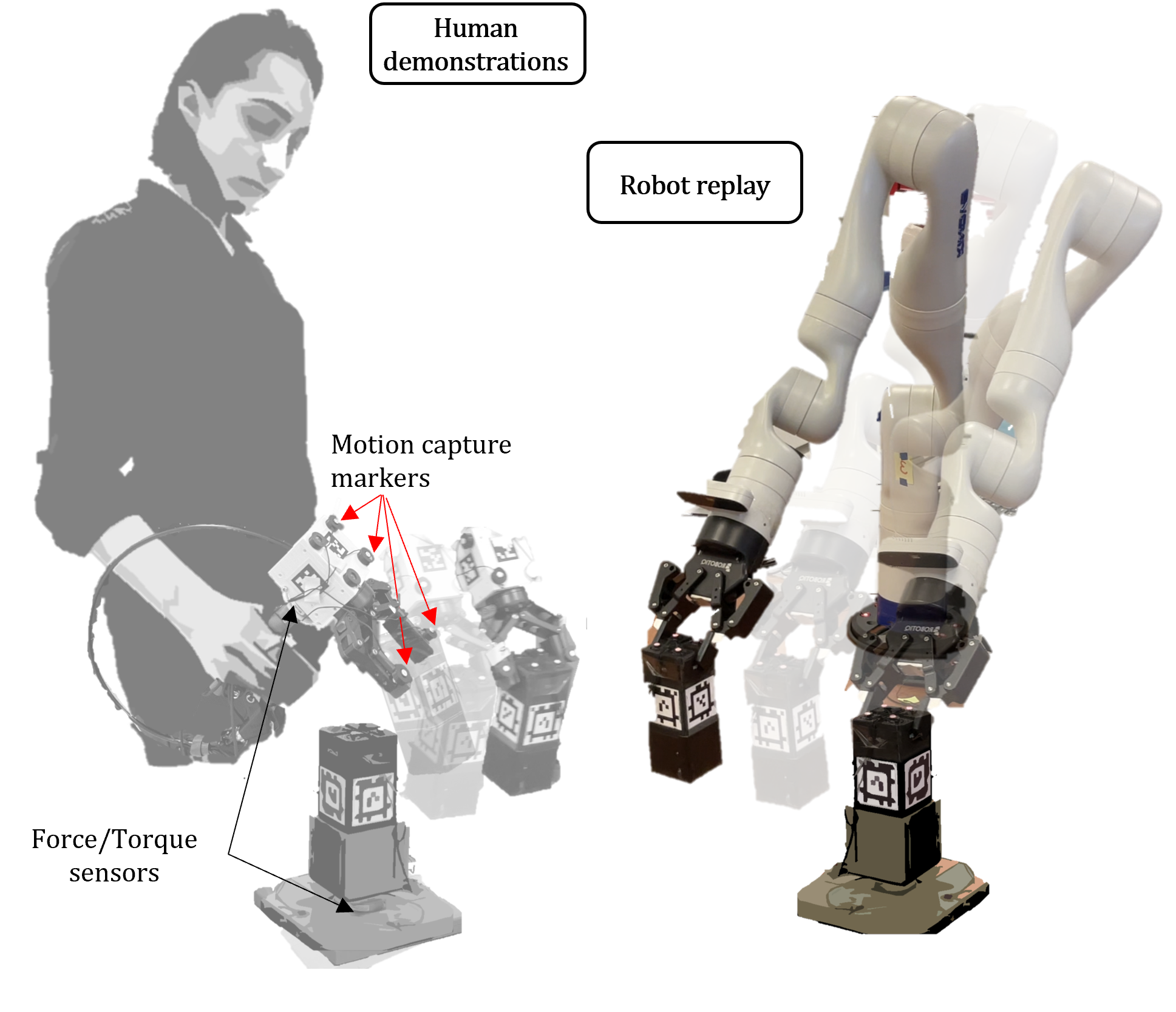}}
       \subfloat[Sensorized tool design\label{FIG:tool_design}]{%
        \includegraphics[width=0.45\textwidth]{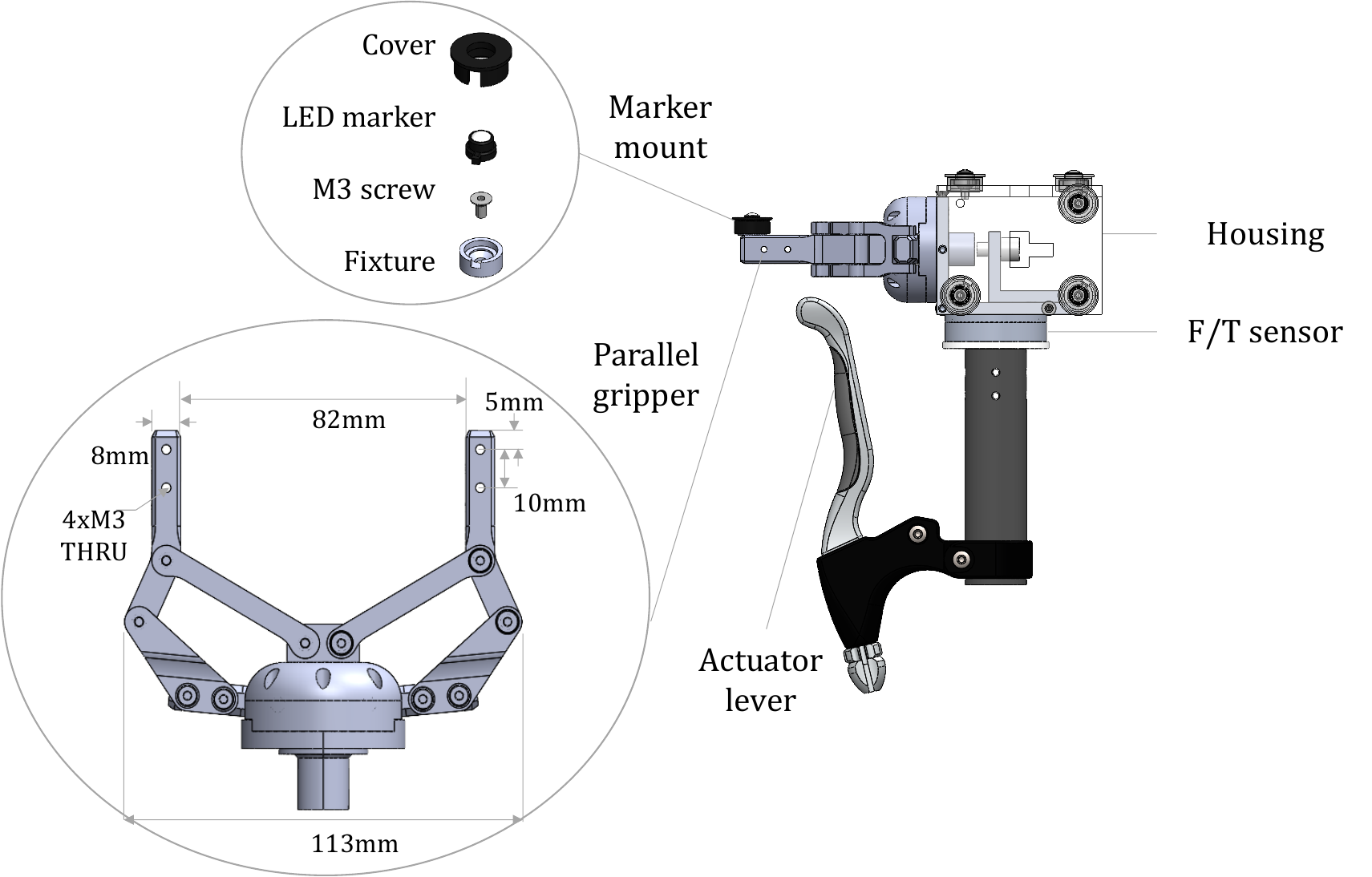}}
  \caption{a) The tool is the only common aspect in the task. b) We choose off-the-shelf gripper and bicycle brake handle for the actuator lever to assemble the tool.}
  \label{FIG:intro_pic}
\end{figure}

The choice of deploying learning from demonstration (LfD) approaches \cite{Calinon2016} is well motivated by the need to make robot programming easy and accessible. Conventional choices in LfD include (i) directly recording humans by motion capture \cite{Artemiadis2010} (ii) Kinesthetic teaching \cite{Kormushev2011} and (iii) immersive teleoperation \cite{Kana2023}. Both teleoperation and kinesthetic teaching can be expensive and need to be safe for human interaction. The morphological differences \cite{Bicchi2000} between humans and robots make it hard to record human motion directly, e.g., a human hand with fingers is hardly comparable to a robot parallel jaw gripper. Anthropomorphic machine hands via sensorized gloves address this \cite{Perret2018, Almeida2019}. In this work, we propose instead to provide the human with a sensorized gripper (Fig. \ref{FIG:intro_pic}) to leverage on the existing skills and dexterity of humans.

In recent years, there has been an increasing trend in developing hand-held sensorized grippers \cite{Song2020}-\cite{Chi2024}. Vision based grasping datasets \cite{Song2020}-\cite{Shafiullah2023} have been produced with focus on manipulation policies obtained from object-centered observations. Similar to these works, we derive a \textit{Cartesian} description of the task. Although visuo-tactile sensing \cite{Suh2022} was recently incorporated in the universal manipulation interface \cite{Chi2024}, to the best of authors' knowledge, wrench sensing is missing in hand-held sensorized tool literature. Task wrenches can help the robot to adapt its motion in the case of failures, especially in the case of occlusions and visual uncertainty. In this work, we compare the haptics of human demonstration with the robot replay of the same task, and identify the possibility of robot trajectory adaptation as a future scope. For this reason, we choose to use impedance control to drive the robot, as opposed to standard position control.

The scope of this work is only rigid body kinematics for now, in order to methodically develop this sensorized gripper platform. Once demonstration statistics are extracted, the robot will be driven at all joints. This requires the resolving of joint \textit{redundancy} both for the robot joint space and its base positioning. \HT{The main goal is to quantify the success rate during the task repetition by the robot and the difference between task forces recorded during demonstration and actual robot replay, i.e., the \textit{haptic mismatch}.}

\section{Cartesian approach to robot motion generation}
\label{sec:motion-generation}

A robot's coordinates usually are defined w.r.t the joint angles $\V q \in \mathbb{R}^D$, while the task variables live in a Cartesian space. This poses the classical problem of determining robot kinematics by (i) computing joint space solution and (ii) determining an optimal location for the robot base. The tool in this paper is a gripper, whose pose is defined as $(\V R_G, \ \V p_G) \in SE(3)$. The distance between the gripper fingers is $d_G \in \mathbb{R}^+$.

\begin{figure}
        \centering
        \subfloat[Optimising robot base\label{3a}]{%
       \includegraphics[width=0.3\textwidth]{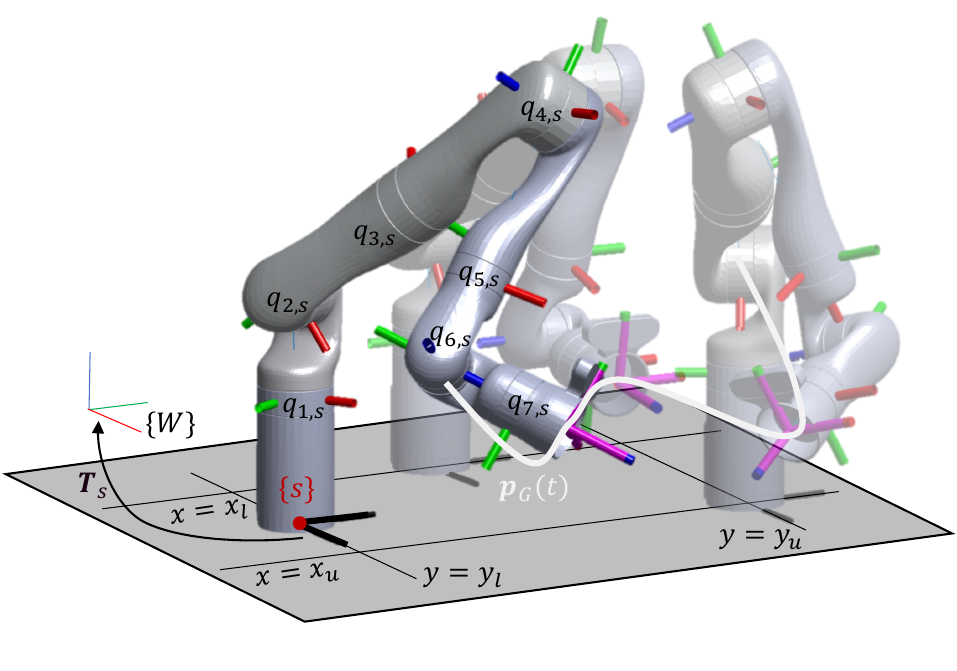}}
        \subfloat[Joint space solution\label{FIG:w_error}]{%
       \includegraphics[width=0.3\textwidth]{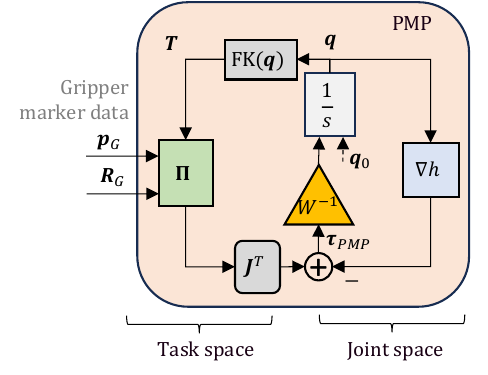}}
  \subfloat[Computing manipulability\label{3b}]{%
        \includegraphics[width=0.4\textwidth]{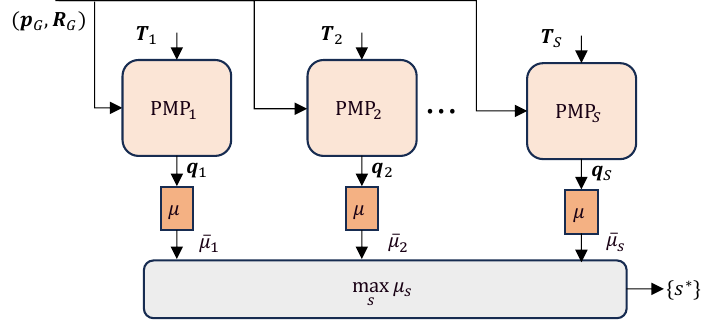}}
  \caption{a) For robot base frame scenario $\{s\}$ varying in a plane, robot manipulability is evaluated. b) Robot joint state is computed from gripper marker data using first order dynamics. c) The optimal robot base frame is determined after optimising the integral of manipulability across all scenarios.}
  \label{FIG:robot_base}
\end{figure}

The sensorized gripper motion is recorded by the motion capture system, as a time indexed data-point $\V m_i(t)$ for marker $i$. Using SE(3) filtering \cite{GiaHoang2020}, we compute task space position $\V p_G(t)$ and orientation $\V R_G(t) \in SO(3)$ of the gripper frame from this marker data. The task space command $\V \Pi$ drives the robot gripper pose $\V T$ (computed using the forward kinematics block $FK$) to track the demonstrated sensorized gripper pose $(\V p_G, \V R_G)$ with an initial configuration $\V q_0$ and the joint torque command $\V \tau_{PMP}$. We implement first order system dynamics to compute the robot state $(\V q, \dot{\V q})$ as outputs (Fig. \ref{FIG:w_error}). Thus, there are two components in $\V \Pi$, the linear error and the rotational error with their corresponding stiffness $k_l$ and $k_r$:
\begin{equation}
    \V \Pi =  \MAT{k_l\V R^T(\V p_G - \V p)\\ 
    k_r\log_\vee(\V R^T \V R_G)}
\end{equation}
where, $\MAT{\V R & \V p\\ \V 0^T & 1} = FK(\V q)$ and $\log_\vee$ is the $SO(3)$ logarithm. We choose the cost function $h$, as one of the possible distances from robot singularity \cite{Chiaverini2008} with the aim of maximising manipulability during the robot motion:
\begin{equation}
    h = -\text{det}(\V J\V J^T)
\end{equation}
Using the identity for the derivative of a matrix determinant in \cite{Petersen2008}, we can compute the gradient $\nabla_{\V q} h$ of the cost as:
\begin{equation}
    \nabla_{\V q} h = \left[\frac{\partial h}{\partial q_1} \dots \frac{\partial h}{\partial q_7}\right]^T
    \label{eq:grad_h}
\end{equation}
The individual gradient terms are computed using:
\begin{equation}
    \frac{\partial h}{\partial q_i} = -\text{det}(\V J \V J^T)\cdot\text{tr}\left((\V J \V J^T)^{-1}\left(\frac{\partial \V J}{\partial q_i} \V J^T + \V J\frac{\partial \V J^T}{\partial q_i}\right)\right) \nonumber
\end{equation}

\subsection{Determining optimal robot location}
\label{subsec:robot-base}

Given a sensorized tool demonstration, it is not trivial to identify the robot base location, since the resultant robot joint space configurations for tool trajectory tracking, might not be optimal in terms of the robot's manipulability. We optimise manipulability and compute it as a functional of the gripper trajectory given by marker data $(\V p_G(t), \V R_G(t))$ (Fig. \ref{FIG:robot_base}). For each scenario $s$, we compute the robot joint trajectory $\V q_s(t) = [q_{1,s} \dots q_{7,s}]^T$ with the PMP block forward kinematics adjusted as if the robot is situated at frame $\{s\}$:
\begin{equation}
    \MAT{\V R & \V p\\ \V 0_{1\times 3} & 1} = \V T_s FK(\V q)
\end{equation}

The cost integral, as an \textit{average manipulability} across the entire volume of the gripper trajectory is computed as:
\begin{align}
 \label{eq:manipulability}
    \bar{\mu} &=\frac{1}{t_1-t_0}\int_{t_0}^{t_1} \text{det}(J(\V q(t))J^T(\V q(t)))\,dt.
\end{align}
where, $J$ is the robot Jacobian. The robot base is varied in position with limits $x \in [x_l \dots x_u]$, $y \in [y_l \dots y_u]$, and for every scenario $s$, we compute the robot manipulability $\mu_s$. The optimal robot base pose $\{s^*\}$ is then selected which maximises this quantity.

\section{From human demonstrations to robot replay}
\label{sec:demo2robot}
In this work, we used the sensorized gripper to perform a square peg-in-hole task. Kinematically, the two frames that describe the task completely are (i) the gripper frame $\{G\}$ and (ii) the box frame $\{B\}$. The demonstrations were performed by a human operator using a metronome to \textit{pace} oneself to allow for consistent motions across trials, to capture some repeatability across time. The gripper frame information $(\V p_G(t_j), \V R_G(t_j))$ for each trial $j$ was computed using the SE(3) filter detailed in \cite{GiaHoang2020}. Additionally, two markers were placed on the gripper fingers to measure the distance $d_G(t_j)$ between them to indicate when the gripper was closed and opened by the human operator. After repeated trials, this information was used to perform Gaussian Mixture Regression to compute the \textit{average} demonstration trajectory in the task space to command the robot to follow.

\subsection{Gaussian Mixture Regression (GMR) of demonstration}

\begin{figure}
\vspace{0.5cm}
\centering
  \includegraphics[width=\textwidth]{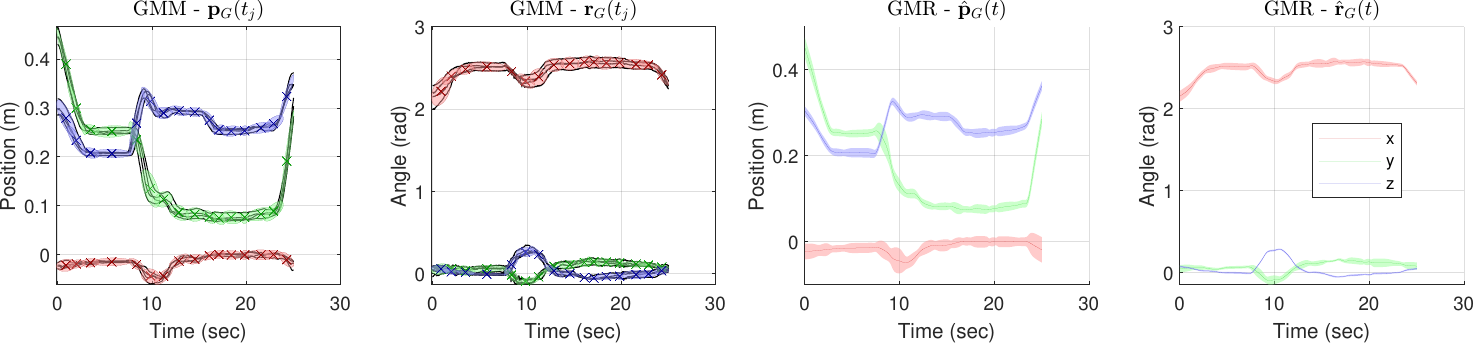}
  \caption{Statistics of human demonstration using sensorized gripper.}
  \label{FIG:GMR_result}
\end{figure}

The notation for GMM/GMR is borrowed from \cite{Calinon2016}. While accounting for the different phases of the task, but also to minimise the mixture's dimensionality we select the number of clusters $K = 16$. The state vector $\V \xi = \{\xi_t, \V{\xi}_s\}$ is composed of a temporal state $\xi_t$ and a spatial state $\V \xi_s$ in a Gaussian Mixture model. In this work, for each trial $j$ we define a dataset $\V {\xi}_j$, with $\V {\xi}_{t,j} = t_j$ and $\V{\xi}_{s,j} = [\V p_G^T(t_j) \  \V r_G^T(t_j) \ d_G(t_j)]^T \in \mathbb{R}^7$, where,
\begin{equation}
    \V r_G(t_j) = \log_\vee(\V R_G(t_j))
\end{equation}

The complete dataset to be passed to GMM is composed of all the samples across different trials, concatenated as $\V \xi = \{\V \xi_j\}_{j=1}^{N} \in \mathbb{R}^{D \times N}$, with $D = 7$. The resultant conditional probability $\V{\hat{\xi}}_{s,t} = [\hat{\V p}_G(t) \  \hat{\V r}_G(t) \ \hat{d}_G(t)]^T$ after the GMR is performed is shown in Fig. \ref{FIG:GMR_result} and is used as the input to the robot motion generation procedure detailed in section \ref{sec:motion-generation} to compute the joint space solution $\hat{\V q}(t)$. We omit plotting $\hat{d}_G(t)$ for brevity.

\subsection{Robot replay}
\label{subsec:robot_replay}
\begin{figure}[h]
        \centering
  \subfloat[Optimising for robot base\label{FIG:optimal_robot_location}]{%
       \includegraphics[width=0.4\textwidth]{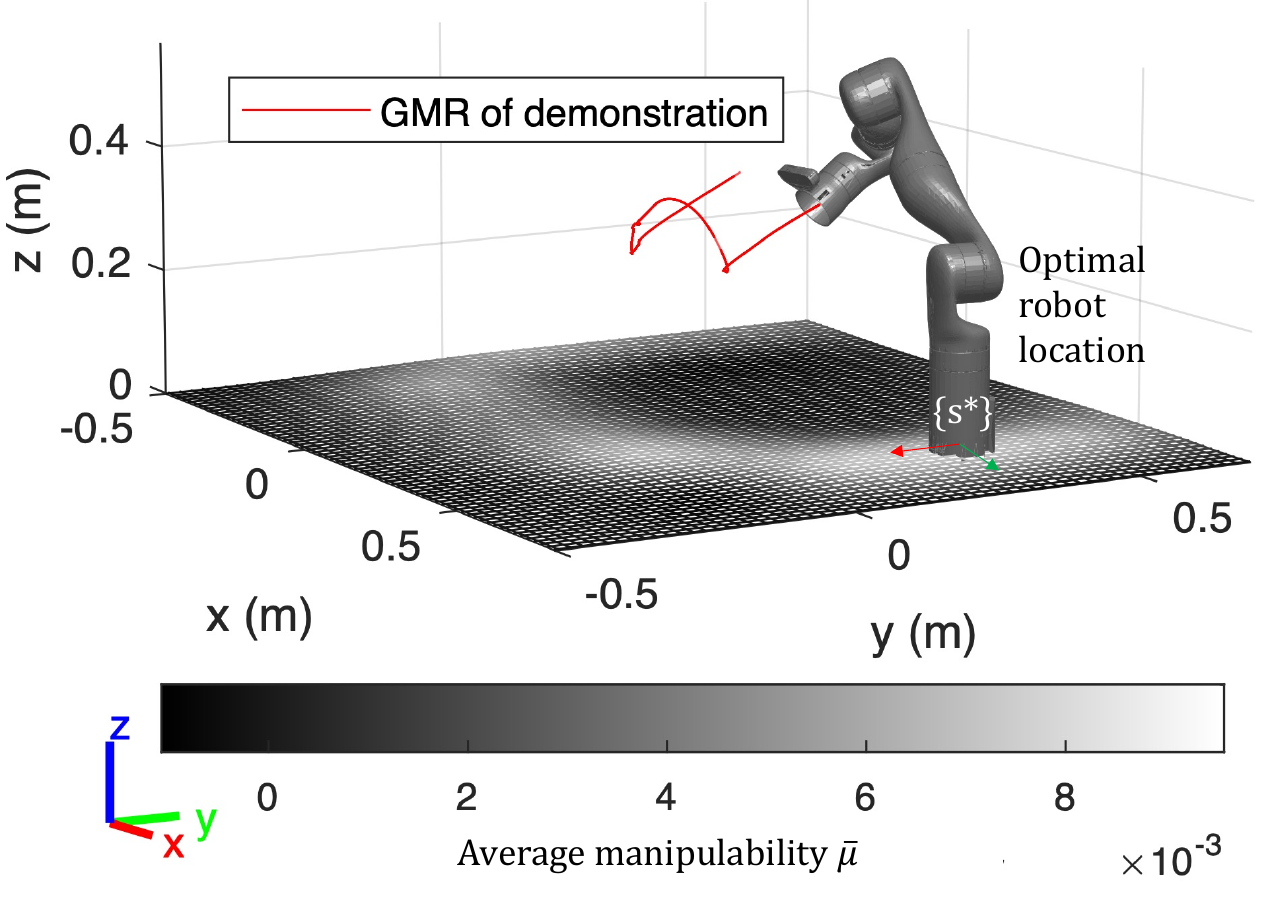}}
  \subfloat[Impedance control\label{FIG:impedance_control_v1}]{%
        \includegraphics[width=0.55\textwidth]{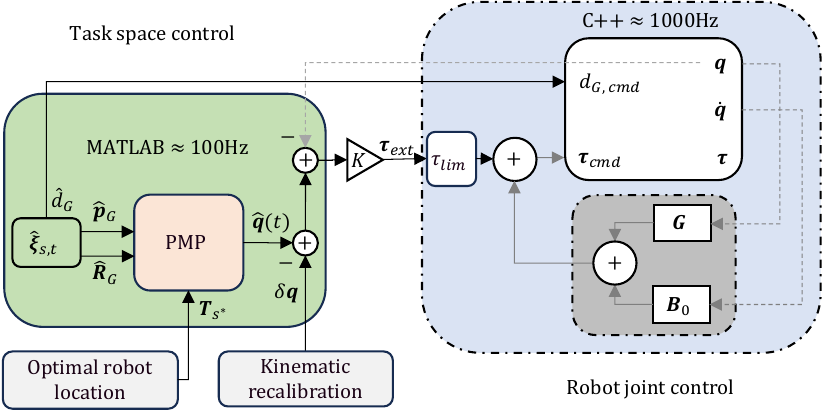}}
  \caption{a) Optimal robot base location computed over a large region surrounding the demonstration. b) Impedance control.}
  \label{FIG:robot_experiment}
\end{figure}

We use a 7DoF Kinova Gen3 robot to replay the human demonstration using the sensorized gripper. The GMR of the human demonstration $\V{\hat{\xi}}_{s,t}$ is used to compute the optimal robot location using the steps detailed in subsection \ref{subsec:robot-base}. We evaluated the \textit{average manipulability} (see eq. \ref{eq:manipulability}), for each scenario $\V T_s$ with $\V p_s = [p_{s,x} \ p_{s,y} \ p_{s,z}]^T$ varying over 100 points uniformly distributed over $p_{s,x}, p_{s,y}\in[-1 \ 1]$, with the optimal result shown in Fig. \ref{FIG:optimal_robot_location}.

\begin{figure}
        \centering
  \subfloat[Robot setup\label{FIG:successful_assembly}]{%
       \includegraphics[width=0.25\textwidth]{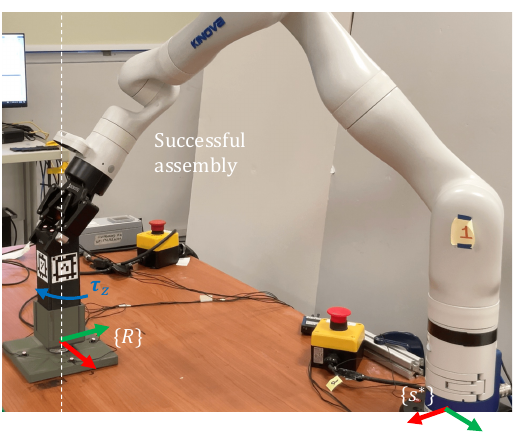}} 
  \subfloat[Robot replay\label{FIG:robot-replay}]{%
        \includegraphics[width=0.25\textwidth]{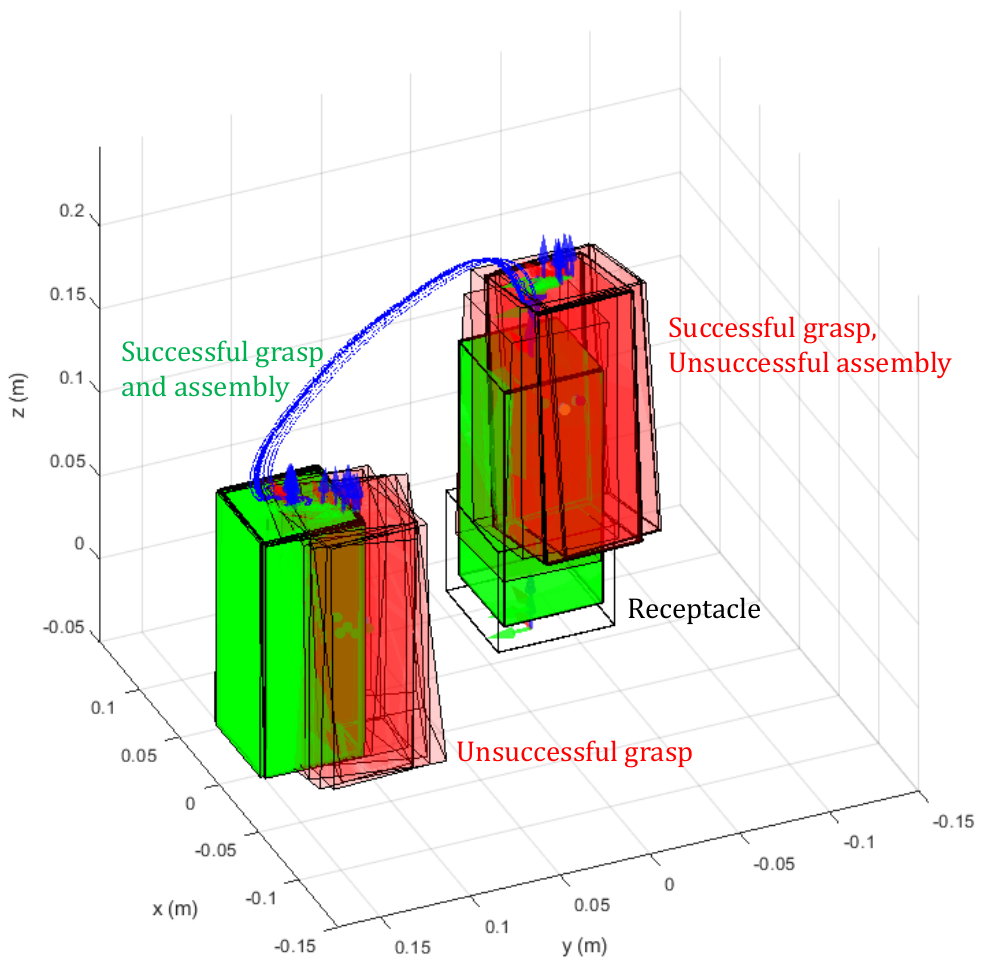}}
        \subfloat[Haptic mismatch\label{FIG:haptic_mismatch}]{%
        \includegraphics[width=0.5\textwidth]{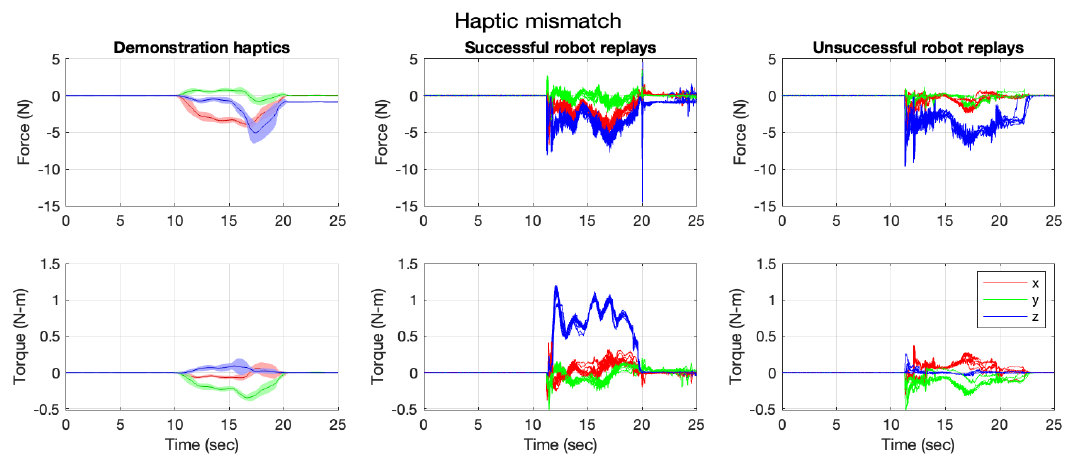}}
  \caption{(a) Task haptics were sensed at the receptacle frame $\{R\}$, (b) In this work, we deliberately vary the initial location of the box to record the resulting \textit{haptic mismatch} in case of in-hand pose uncertainty, (c) Haptic mismatch, i.e., the difference in demonstration forces and robot forces as measured in the receptacle frame $\{R\}$.}
  \label{FIG:results}
\end{figure}

Before the replay of the demonstration on the robot, we re-calibrate \cite{Kana2022} the joint kinematics of the robot to improve accuracy. The optimal robot location is used to position the robot and the robot pose w.r.t the receptacle frame is re-estimated using motion capture data. The resultant joint space solution $\hat{\V q}(t)$ is coupled with the joint offset results $\delta \V q$ from kinematic recalibration to perform impedance control (Fig. \ref{FIG:impedance_control_v1}) on the robot with stiffness $K$. The computed torque command $\V \tau_{ext}$ is passed through a filter to check the torque limits $\tau_{lim}$ to ensure safety. Simultaneously, the gripper position data $\hat{d}_G(t)$ is relayed to the robot. A total of 30 trials were carried out to test the statistics of the robot replay \HT{(Fig. \ref{FIG:successful_assembly})} and document the cases where success/failure occurred for different box initial locations (Fig. \ref{FIG:robot-replay}). \HT{We deliberately place the box in slightly different locations and orientations to quantify the \textit{haptic mismatch} between demonstrations and robot replays in the case of in-hand pose uncertainty (Fig. \ref{FIG:haptic_mismatch}). A summary video of this work is available online \cite{video}.}

One of the main contributions of this work is the ease of programming the robot using the sensorized gripper. Each trial in the demonstration took approximately 25$sec$. A total of 4 trials, i.e., 100$sec$ of demonstration data was used to produce the results presented in this work. Even though task-parameterised generalisation is a well studied topic \cite{Calinon2016}, \cite{Hu2019} which allows to overcome in-hand pose uncertainty upon calibration, in this work our goal is to quantify the haptic mismatch between demonstrations and robot replays, in cases where the box was slightly displaced from the initial pick-up location. We specifically focus on the torques measured (Fig. \ref{FIG:haptic_mismatch}) at the receptacle frame $\{R\}$. The demonstration haptics from the GMR can be compared to the robot replay haptics. Clearly the successful assembly haptics itself does not match the demonstration. However, it is useful to observe that a high torque in the $z$-direction is indicative of successful assembly, since only after the box has entered the receptacle, could it actuate a high enough rotational torque $\tau_z$ about the vertical (see Fig. \ref{FIG:successful_assembly}).

\section{Conclusion}\label{sec:conclusion}
In this work, a sensorized gripper built using off-the-shelf parts was used to perform a box in box assembly task. Our key contribution is the ease of programming robots, with a total of 100$sec$ of demonstration data used to produce quick robot replay results with successful assembly outcomes. Demonstration statistics were extracted using Gaussian Mixture Models (GMM). The optimal robot location was computed while concurrently determining the joint space solution to track the task demonstration. Finally the robot was commanded in impedance control to repeat the task several times to study success rate. \HT{We quantified the haptic mismatch between human demonstrations and robot replays and identified trajectory adaptation as a future scope.}
\backmatter

\bmhead{Acknowledgements}

This research is supported by the National Research Foundation, Singapore, under the NRF Medium Sized Centre scheme (CARTIN). Any opinions, findings, and conclusions or recommendations expressed in this material are those of the author(s) and do not reflect the views of National Research Foundation, Singapore. The authors would also like to thank Choo Wei Jie for his contributions to the design of the initial prototype of the sensorized gripper.

%% if required, the content of .bbl file can be included here once bbl is generated
%%\input sn-article.bbl

\end{document}